\documentclass[letterpaper, 10 pt, conference]{ieeeconf}  

\IEEEoverridecommandlockouts                              

\RequirePackage{mathtools}  
\RequirePackage{amssymb}    
\RequirePackage{siunitx}    

\RequirePackage{tabularx}   
\RequirePackage{booktabs}   
\RequirePackage{longtable}  
\RequirePackage{multirow}   
\RequirePackage{colortbl}
\RequirePackage{xcolor}

\RequirePackage{graphicx}   
\RequirePackage{float}      
\RequirePackage[labelfont=bf,justification=centering, footnotesize]{caption} 
\RequirePackage{subcaption}         
\RequirePackage{algpseudocode}      
\RequirePackage{algorithm}          
\usepackage{hyperref}
\usepackage{soul,xcolor}
\usepackage[normalem]{ulem}
\usepackage[backend=biber, style=numeric-comp, sorting=none]{biblatex}
\AtEveryBibitem{
  \clearfield{doi}
  \clearfield{url}
  \clearfield{issn}
}

\newcommand{\algorithmwithoutend}{%
  \algtext*{EndWhile}%
  \algtext*{EndFor}%
  \algtext*{EndLoop}%
  \algtext*{EndIf}%
  \algtext*{EndProcedure}%
  \algtext*{EndFunction}%
}

\definecolor{lightgray}{gray}{0.92}
\definecolor{lightgreen}{rgb}{0.4019607843137255, 0.7862745098039216, 0.3901960784313726}
\setstcolor{red}

\title{\LARGE \bf
Pushing Through Clutter With Movability Awareness of \\ Blocking Obstacles
}

\author{Joris J. Weeda, Saray Bakker, Gang Chen and Javier Alonso-Mora
\thanks{This project has received funding from the European Union through ERC, INTERACT, under Grant 101041863. Views and opinions expressed are however those of the author(s) only and do not necessarily reflect those of the European Union. Neither the European Union nor the granting authority can be held responsible for them.} 
\thanks{The authors are with the Cognitive Robotics Department, TU Delft,
2628 CD Delft, The Netherlands 
        {\tt\small \{jjweeda, s.bakker-7, g.chen-5}\}@tudelft.nl}%
}

\begin{document}

\maketitle
\thispagestyle{empty}
\pagestyle{empty}

\begin{abstract}
Navigation Among Movable Obstacles (NAMO) poses a challenge for traditional path-planning methods when obstacles block the path, requiring push actions to reach the goal. 
We propose a framework that enables movability-aware planning to overcome this challenge without relying on explicit obstacle placement.
Our framework integrates a global Semantic Visibility Graph and a local Model Predictive Path Integral (SVG-MPPI) approach to efficiently sample rollouts, taking into account the continuous range of obstacle movability. A physics engine is adopted to simulate the interaction result of the rollouts with the environment, and generate trajectories that minimize contact force.
In qualitative and quantitative experiments, SVG-MPPI outperforms the existing paradigm that uses only binary movability for planning, achieving higher success rates with reduced cumulative contact forces.
Our code is available at: \href{https://github.com/tud-amr/SVG-MPPI}{https://github.com/tud-amr/SVG-MPPI}

\end{abstract}

\begin{keywords}
    Motion and Path Planning, 
    Collision Avoidance,
    Integrated Planning and Control
\end{keywords}


\section{INTRODUCTION}
\label{section: introduction}
A fundamental ability of autonomous robots is to navigate towards a goal while avoiding collisions along the way~\cite{Xiao2020}. 
However, in complex and cluttered environments, such as domestic settings where obstacles like chairs and boxes may obstruct the path to the goal, finding collision-free paths often becomes impractical.
In such cases, traditional navigation methods often fail and Navigation Amongst Movable Obstacles~(NAMO) becomes essential.

NAMO involves actively interacting with the environment by relocating obstacles to clear a path, enabling successful navigation to the target. 
The concept was initially explored by \citeauthor{Reif1985} \cite{Reif1985}, and later formally introduced as a category of research by \citeauthor{Stilman2004} \cite{Stilman2004}. These works demonstrated that solving NAMO in a two-dimensional workspace with movable obstacles is NP-hard, and NP-complete in simplified cases with unit-square obstacles, due to the large search space and the numerous possible configurations \cite{Reif1985, Demaine2000}. As a result, most solutions are approximations rather than exact solutions, and practical applications remain limited~\cite{Ellis2022}. Research often focuses on simplified versions of the problem, such as $LP_1$, a linear program where a single obstacle blocks the path, or $LP_2$, which involves two obstacles \cite{Stilman2004, Renault2020}.

\begin{figure}[tb!]
    \includegraphics[width=0.49\textwidth]{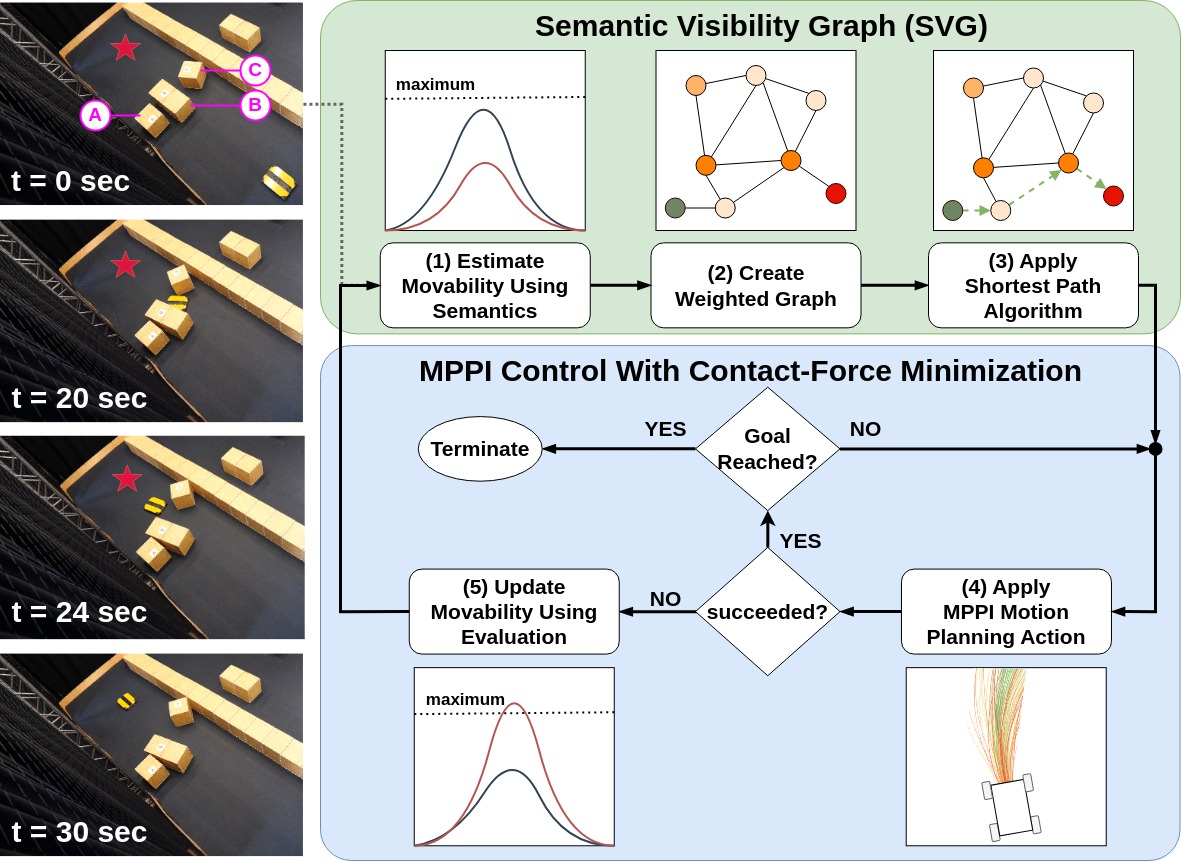}
    \caption{An overview of the proposed SVG-MPPI architecture where the SVG provides a weighted graph with efficient node placement around movable obstacles along which a lowest-effort path can be found. The generated set of waypoints guides the MPPI control strategy to efficiently sample rollouts around movable obstacles. If during interaction an obstacle is considered non-movable, the movability estimation gets updated and the path is replanned. Snapshots of a real-world example are shown on the left where the red star indicates the goal location and the masses of the obstacles are (A): 25 kg, (B): 20 kg, (C): 5 kg.}
    \label{fig:overview_of_svg_mppi}
    \vspace{-4.5mm}
\end{figure}

Existing NAMO approaches \cite{Demaine2000, Stilman2004, Stilman2010, Nieuwenhuisen2006, Mueggler2014, Castaman2016, Moghaddam2016, Meng2018, Ellis2022} and sequential nonprehensile manipulation strategies~\cite{Muguira2023Visibility, vieira2022persistent} consider only binary movability information, i.e., whether obstacles are movable or not, neglecting the varying levels of effort required to move different obstacles and thus lacking efficiency.
Furthermore, these works rely on task planners to decide when the robot should move without displacing an obstacle (transit) or when it should displace an obstacle (transfer) and where to place the obstacle. 
However, obstacle placement planning itself is a difficult problem and these works typically function only in simplified scenarios, such as the $LP_1$ and $LP_2$ scenarios. How to realize efficient NAMO in more cluttered and random environments remains an open problem.

This work incorporates continuous movability information into the NAMO-solving process, improving efficient navigation around obstacles with contact force minimization while moving to the goal. 
Moreover, inspired by the framework in~\cite{Ellis2022}, we use a physics engine to model state transitions of the obstacles in the environment. We introduce a framework that can rapidly simulate the interaction outcomes of each rollout first before selecting the best one, without the need for explicit obstacle placement planning, in contrast to~\cite{Ellis2022}. These rollouts are planned using an SVG-MPPI approach, which tackles global planning with movability-aware node placement and local planning with the consideration of contact force minimization using a Model Predictive Path Integral (MPPI) control strategy.
Fig. \ref{fig:overview_of_svg_mppi} shows the framework of our system and snapshots of a real-world experiment.

The main contributions of this work are:
\begin{itemize}
    \item \textit{Integration of Continuous Quantified Movability:} We consider obstacle movability on a continuous scale rather than as a binary property, leading to more efficient planning in cluttered environments.

    \item \textit{Contact-Force Minimization in MPPI control:} Our approach minimizes contact forces between the robot and the environment using the movability data and a physics engine, reducing the effort required to move objects.

    \item \textit{Elimination of Explicit Obstacle Placement:} We eliminate the requirement for explicit obstacle relocation, enabling applicability in more cluttered random environmental setups.
\end{itemize}

\section{PRELIMINARIES}
\label{chapter:preliminaries}

This section presents the key concepts underlying our approach. We begin by defining the necessary notations, followed by an overview of the Visibility Graph (VG) method which provides the basis for the SVG. Lastly, we describe the MPPI control strategy, which is integrated with a physics engine to model obstacle interactions.

\subsection{Notation}
\label{sec: notation}

In this paper, a graph is represented as \( G = (V, E) \), where \( V = \{v_1, v_2, \dots, v_{n}\} \) is the set of nodes with $n$ the number of nodes, and \( E \) the edges. Each obstacle is described as a polygon \( O_i \), with mass \( m_i \). The distance between a node \( v_i \) and an obstacle is denoted as \( d_i \), and the robot's safety margin is given by \( r \). \textit{Free-space nodes}, \( V_{\text{free}} \), originate from the visibility graph and do not require manipulation, while \textit{passage nodes}, \( V_{\text{passage}} \), involve obstacle manipulation. 
The robot’s state is denoted \( \mathbf{x}_t \), its velocity \( \mathbf{\dot{x}}_t \), and control inputs are given by \( \mathbf{v}_t \).
A set of waypoints is represented by \( \mathcal{P} = \{\mathbf{p_1}, \mathbf{p_2}, ..., \mathbf{p}_{n_{p}} \} \) with $n_{p}$ the number of positions $\mathbf{p}$.

\subsection{Visibility Graphs (VG)} \label{sec: VG}

\begin{figure}[tb!]
    \centering
    \begin{minipage}{0.145\textwidth}
        \centering
        \includegraphics[width=1.1\textwidth]{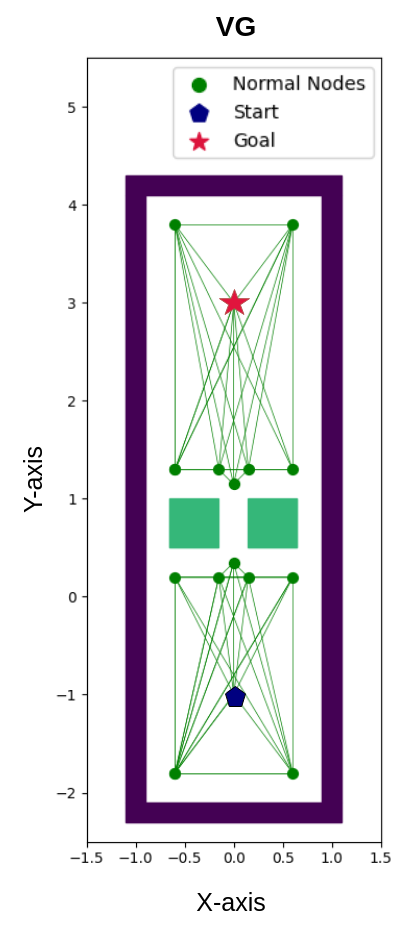}
        \caption*{\hspace{14mm} (a)}
    \end{minipage}\hfill
    \begin{minipage}{0.145\textwidth}
        \centering
        \includegraphics[width=1.1\textwidth]{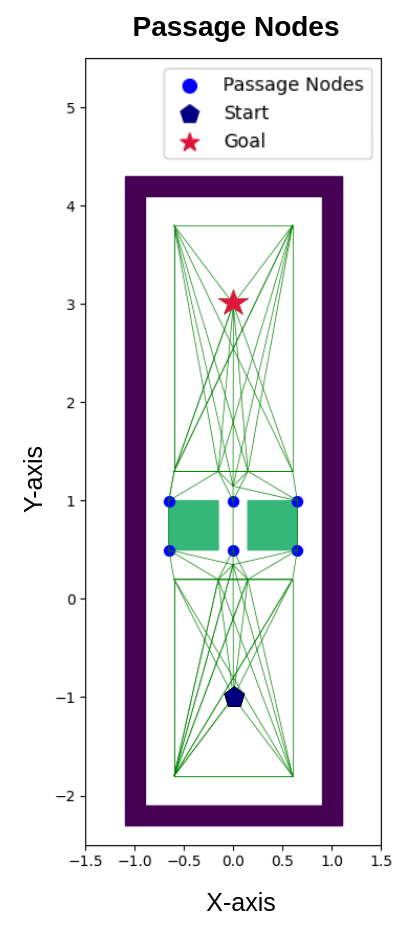}
        \caption*{\hspace{14mm} (b)}
    \end{minipage}\hfill
    \begin{minipage}{0.191\textwidth}
        \centering
        \includegraphics[width=1.1\textwidth]{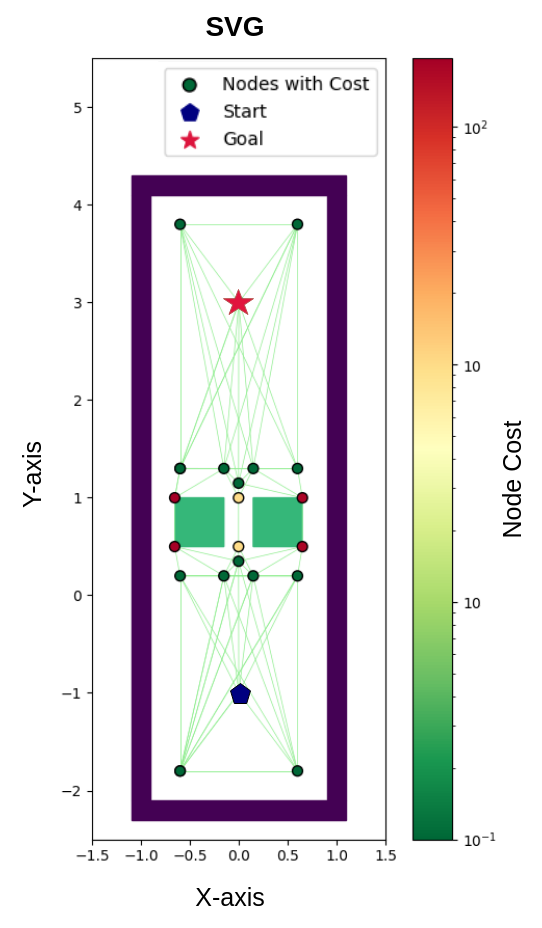}
        \caption*{\hspace{14mm} (c)}
    \end{minipage}
    \caption{Comparison of different visibility graph strategies: (a) The normal Visibility Graph (VG), (b) A VG enhanced with Passage Nodes, and (c) A Semantic Visibility Graph (SVG) where node costs reflect obstacle manipulation effort.}
    \label{fig:simple_svg_graph}
    \vspace{-3mm}
\end{figure}

VG is a key tool in path planning, representing direct line-of-sight connections between significant points in an environment \cite{Berg2000}. These connections, or edges, create a network that allows algorithms such as Dijkstra’s \cite{Dijkstra1959} or A* \cite{Nilsson1968} to compute optimal paths between start and goal nodes. VG construction involves identifying key points, typically at obstacle corners, and connecting them with edges wherever visibility is unobstructed. To ensure safe navigation, obstacle boundaries are inflated by a margin \( r \), resulting in a graph \( G = (V_{\text{free}}, E) \) that facilitates efficient path planning, particularly when obstacles are static. Fig.~\ref{fig:simple_svg_graph}a demonstrates a basic visibility graph applied to a straightforward environment, with inflated obstacles and points placed in the remaining free space. However, Fig.~\ref{fig:simple_svg_graph}a also highlights a limitation of VGs: when the goal is blocked by movable obstacles, certain locations are not mutually visible, creating gaps in the visibility graph. Consequently, VG alone becomes insufficient for generating a path to the goal in such scenarios.

\subsection{Model Predictive Path Integral (MPPI)} 
\label{sec: MPPI}

MPPI is a sampling-based control method used for navigating dynamic environments by generating multiple control input sequences and evaluating them using a cost function~\cite{Williams2017}. 
MPPI involves sampling a total of \( K \) input sequences \( V_k\) that generate the state trajectories \( Q_k, \ \forall k\in K\), often referred to as rollouts.
Each rollout \( Q_k \) represents a series of state vectors \( \mathbf{x}_{t,k} \), meaning that each rollout \( k \) consists of prediction steps \( t \) over a total horizon of \( T \). Using a cost function \( C_k \), each rollout is evaluated where rollouts that violate the constraints are disregarded. A weighted sum of all rollouts determines the next action for the robot. In the work by \citeauthor{Pezzato2023}, MPPI is integrated with IsaacGym \cite{Makoviychuck2021}, a physics engine that simulates the robot’s environment in parallel using GPU power, allowing for rapid computation of multiple trajectories and the elimination of an explicit model of obstacle interactions~\cite{Pezzato2023}.

\section{METHOD}
\label{section: method}

This section introduces SVG-MPPI, a framework for navigating cluttered environments without explicit object placement. It combines the Semantic Visibility Graph (SVG) for path planning and MPPI with contact force minimization for robot control (Fig.~\ref{fig:overview_of_svg_mppi}). Sec.~\ref{sec: SVG} details the SVG algorithm, which adds passage nodes to connect disconnected regions. Sec.~\ref{sec: mppi with contact-force-minimization} explains the MPPI control strategy, using real-time feedback from IsaacGym to minimize contact forces. Finally, Sec.~\ref{sec: movability evaluation} covers movability evaluation for online replanning based on obstacle interactions.

\subsection{Semantic Visibility Graph (SVG)}
\label{sec: SVG}

The SVG extends the traditional VG by incorporating continuous movability, allowing the placement of \textit{passage nodes} near movable obstacles while accounting for the push actions required to access them. These nodes connect previously disconnected areas, creating new pathways for the robot, as shown in Fig.~\ref{fig:simple_svg_graph}b-c. In the initial phase (lines 1-11 of Algorithm \ref{alg:svg}), the process follows the conventional VG approach \cite{Berg2000}. In the second phase (lines 12-19 of Algorithm \ref{alg:svg}), passage nodes are added near movable obstacl
es where the space between pairs of obstacles is too narrow for the robot to pass. These passage nodes are integrated into the weighted visibility graph like any other node. However, accessing these nodes requires the robot to manipulate the obstacle as the safe distance margin is breached.

\begin{algorithm}[b!]
    \caption{Semantic Visibility Graph}
    \label{alg:svg}

    \algorithmwithoutend

    \footnotesize
    \begin{algorithmic}[1]
        \color{darkgray}
        \State \textbf{Input:} A set \( O \) of disjoint polygonal obstacles.
        \State \textbf{Output:} The visibility graph \( G(V, E) \).
        \vspace{1mm}

        \State Initialize \( V \gets \emptyset \); \( E \gets \emptyset \)
        \vspace{1mm}
        \For{$i \gets 0$ to $|O| - 1$}
            \State Compute \( P_i \gets \text{vertices of inflated polygon } O_i \text{ with safety margin } r \)
            \For{$j \gets 0$ to $|P_i| - 1$}
                \State Select \( v \gets P_i[j] \)
                \State Compute \( W \gets \text{visibleVertices}(v, O) \)
                \For{each vertex \( w \in W \)}
                    \State Update \( E \gets E \cup \{(v, w)\} \)
                \State Update \( V \gets V \cup \{v\} \)
                \EndFor
            \EndFor
        \EndFor

        \vspace{1mm}
        
        \color{black}
        \For{each pair \( (P_i, P_j) \) where \( i \neq j \)}
            \If{distance between \( P_i \) and \( P_j \) is less than \( 2r \)}
                \If{either \( \text{mass}(P_i) \leq \text{max mass} \) or \( \text{mass}(P_j) \leq \text{max mass} \)}
                    \State Compute \( p \gets \text{passageNodes}(P_i, P_j) \)
                    \State Compute \( W \gets \text{visibleVertices}(p, O) \)
                    \For{each vertex \( w \in W \)}
                        \State Update \( E \gets E \cup \{(p, w)\} \)
                    \State Update \( V \gets V \cup \{p\} \)
                    \EndFor
                \EndIf
            \EndIf
        \EndFor
        \color{black}
        \vspace{1mm}
        
        \State \Return \( G = (V, E) \)
    \end{algorithmic}
    \normalsize
\end{algorithm}

Movability is treated as a continuous variable, based on the mass \( m_i \) of each obstacle. Objects with masses below a threshold are considered movable, with lighter objects preferred for manipulation. Movability, defined by mass, can be estimated using perception-based machine learning models such as Image2Mass \cite{Standley2017}, or derived from obstacle interactions \cite{Wu2015, Aguilera2021, Dutta2023, Yuan2023}. However, the process of deriving mass from environmental data is beyond the scope of this work, meaning mass distributions are assumed to be known.

\begin{figure}[tb!]
    \includegraphics[width=0.49\textwidth]{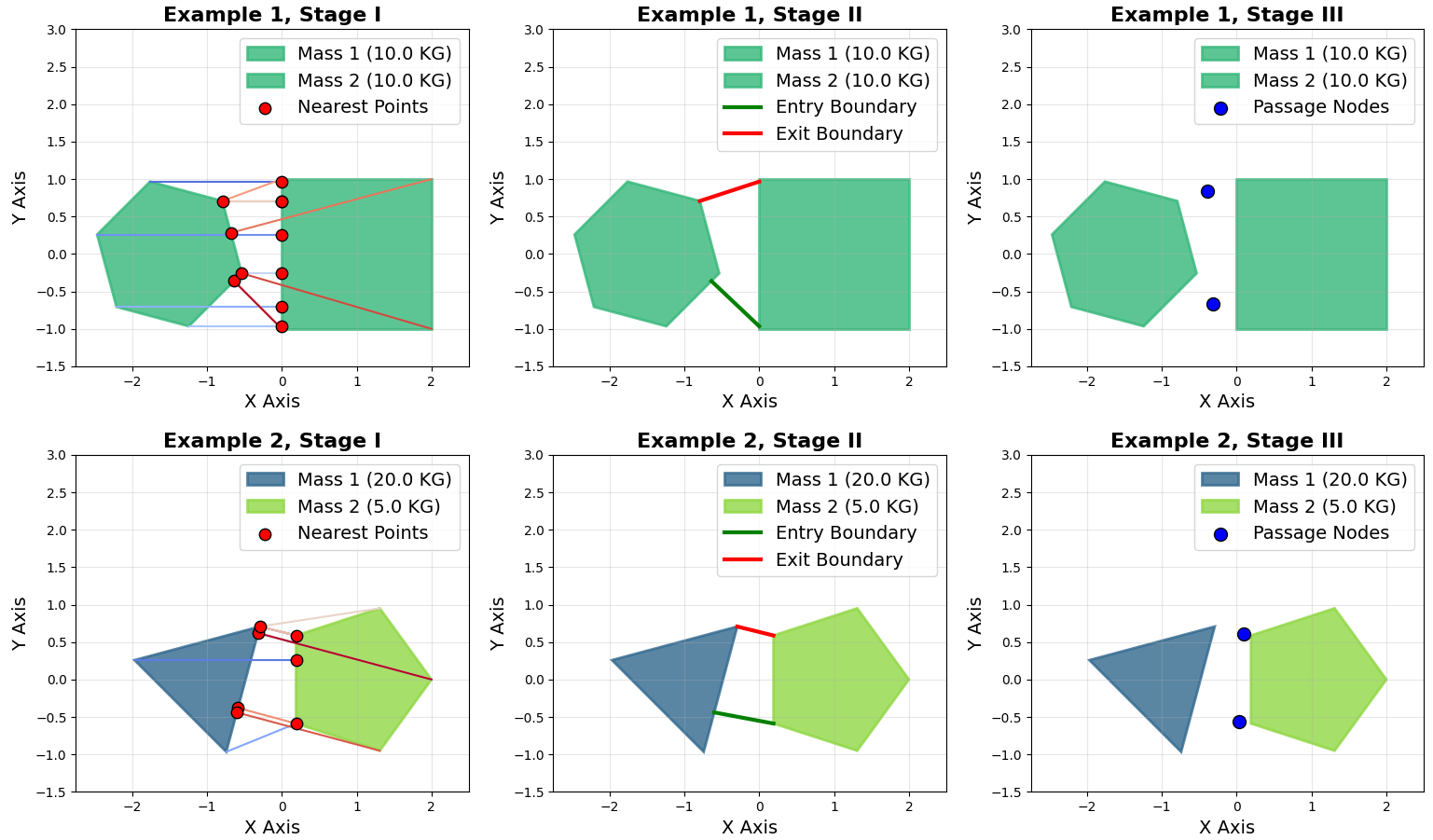}
    \caption{Visualization of passage node construction between two sets of obstacles: (1) Hexagon and square (top row), and (2) Triangle and pentagon (bottom row). Each stage displays the progression from initial shape plotting to the identification of nearest points, boundary construction, and final passage node generation.}
    \label{fig:passage_node_construction}
    \vspace{-4.5mm}
\end{figure}

Passage nodes are created by identifying candidate obstacle pairs. Two obstacles, \( O_i \) and \( O_j \), form a pair if they are close together, at least one is movable, and the gap between them is too narrow for the robot to pass. The boundaries are approximated using convex hulls to ensure accurate node placement for both convex and non-convex shapes. Passage nodes are then created through a three-stage process, as visualized with two examples in Fig.~\ref{fig:passage_node_construction}. 
In Stage I, the nearest points on the edges of the objects are identified that lie on the line-segments connecting the vertices of $O_i$ to the other object $O_j$, to estimate the passage area.
By selecting the closest points within this set,
an area is created for connecting passage nodes to each other and the disconnected regions. In Stage II, entry and exit boundaries are drawn from the convex hull of all the nearest points, defining the line segments where passage nodes will be placed. Finally, In Stage III, passage nodes \( V_{\text{passage}} \) are placed along entry and exit boundaries, treating them interchangeably in an undirected graph. Each passage node is repositioned along the boundary using linear interpolation, shifting it closer to the lighter obstacle to minimize contact with the heavier one. The interpolation factor \( \gamma \) depends on the masses \( m_i \) and \( m_j \) of the obstacles. A passage node \( v_{\text{z}} \in V_{\text{passage}} \), where \( z \in \{1, \dots, n_{\text{passage}} \} \) and \( n_{\text{passage}} \) is the total number of passage nodes, is defined as:

\small
\begin{equation} \label{eq: passage node}
    v_z = v_i + \gamma (v_j - v_i), \qquad \text{where} \ \gamma = m_j 
\left({m_i + m_j}\right)^{-1}.
\end{equation}
\normalsize

In contrast to free space nodes, passage nodes have an additional cost based on the distances from the passage point to nearby obstacles, adjusted by their respective masses. Specifically, for a passage node \( v_z \in V_{\text{passage}}\), the cost \( C_{z} \) is computed as:

\small
\begin{equation} \label{eq:node_cost}
    C_{z} = \max\left(0, 1 - \frac{d_i}{r}\right) \cdot m_i + \max\left(0, 1 - \frac{d_j}{r}\right) \cdot m_j
\end{equation}
\normalsize

The SVG constructs a weighted, undirected graph where edge costs represent travel distances between waypoints, and node costs reflect the effort required to manipulate obstacles (Fig.~\ref{fig:simple_svg_graph}). 
A* \cite{Nilsson1968} is used to compute the optimal path, guided by edge costs for robot movement and manipulation-effort cost of the passage nodes, with the Euclidean distance as a heuristic. 
This ensures that both the distance traveled and the physical manipulation required are accounted for in determining the lowest-cost path. 
Using the set of waypoints, we apply linear interpolation, represented as a first-order polynomial, to ensure uniform spacing between waypoints~\cite{Agarwal1993}. This spacing aids in normalizing the distance cost term for our local control strategy MPPI.

\subsection{MPPI with Contact-Force Minimization}
\label{sec: mppi with contact-force-minimization}

The method introduced in Sec.~\ref{sec: SVG} generates a set of waypoints from the robot's starting position to the goal. In this section, we outline the constraints and objective function of MPPI, which ensure waypoint following while minimizing push actions by reducing contact forces on the robot. 
By integrating MPPI with the IsaacGym physics engine, as described in Sec.~\ref{sec: MPPI}, we extract the robot's contact force tensor and incorporate it into the objective function. 

The objective function is subject to both equality and inequality constraints. The robot’s state transition, \( \mathbf{x}_{t+1} = f(\mathbf{x}_t, \mathbf{v}_t) \), for all \( t \in [0, \dots, T-1] \), is computed using IsaacGym~\cite{Pezzato2023}. The initial state is constrained by the robot’s current pose, \( \mathbf{x}_0 = \mathbf{x}_{\text{init}} \) and inequality constraints ensure that the robot’s state \( \mathbf{x}_t \in \mathcal{X} \) and control inputs \( \mathbf{v}_t \in \mathcal{V} \) remain within the feasible space for all \( t \in [1, \dots, T] \).

MPPI defines state trajectories \( Q_k \) across \( k \in K \) rollouts, each evaluated by a corresponding cost \( C_k \). 
Unlike the implementation in~\cite{Pezzato2023}, we omit the discount factor \( \lambda \) and concentrate most terms within the terminal cost \( C_{\text{term}} \) at the final prediction step rather than distributed across intermediate steps with stage cost \( C_{\text{stage}} \). 
This ensures that MPPI emphasizes long-term planning and allows for deviations from the path at intermediate stages, to avoid unnecessary push actions. The total cost function \( C_{\text{NAMO}} \) for each rollout \( k \) is defined as:

\small
\begin{equation} \label{eq: stage and terminal function}
    C_{\text{NAMO}} = 
    \begin{cases}
        C_{\text{term}}    & ,\text{if} \ t = T \\
        C_{\text{stage}}   & ,\text{otherwise}.
    \end{cases}
\end{equation}
\normalsize

Next, we will discuss the individual cost terms that make up the total cost function. Each cost term, such as \( C_{\text{dist}} \) for distance or \( C_{\text{rot}} \) for rotation, is a weighted version of its corresponding component. For example, the distance cost is expressed as \( C_{\text{dist}} = w_{\text{dist}} \cdot c_{\text{dist}} \), where \( w_{\text{dist}} \) is the associated weight. These weights serve as hyperparameters that balance the different components of the robot's behavior. The stage and terminal costs are structured as follows: 

\small
\begin{subequations} \label{eq: cost function}
    \begin{align}
    C_{\text{term}} &= C_{\text{ctrl}} + C_{\text{dist}} + C_{\text{prog}} + C_{\text{rot}} + C_{\text{force}}, \\
    C_{\text{stage}} &= C_{\text{ctrl}}.
    \end{align}
\end{subequations}
\normalsize

The control cost \( c_{ctrl} \) minimizes deviations between the robot’s predicted velocity \( \mathbf{\dot{x}}_t \), obtained using IsaacGym, and the desired control input \( \mathbf{v}_t \). This cost penalizes jerky movements and encourages smooth trajectories. The weight for the control cost is represented as a diagonal matrix \( W_{\text{ctrl}} \), which assigns different penalties to each velocity component. The control cost is expressed as:

\small
\begin{equation} \label{eq: control cost}
    c_{\text{ctrl}} = \left( \frac{ | \dot{\textbf{x}}_t - \textbf{v}_t | }{\textbf{u}_{\text{max}}} \right)^T W_{\text{ctrl}} \left( \frac{ | \dot{\textbf{x}}_t - \textbf{v}_t | }{\textbf{u}_{\text{max}}} \right).
\end{equation}
\normalsize

The distance cost \( c_{\text{dist}} \) encourages the robot to move toward the next waypoint. First, the set $\mathbf{D}_t = \{d_{1, t}, \dots, d_{n_p, t}\}$ collects the Euclidean distances between the robot’s predicted state \( \mathbf{x}_t \) and all waypoints \( \mathcal{P} \). The closest waypoint $\textbf{p}_{i}$ is identified by finding the index \( i \) that minimizes \( \mathbf{D}_t \). 
The robot then targets the next waypoint \(\mathbf{p}_{i+1}\), where the distance \( d_{i+1,t} \) is normalized by \( \alpha \):

\small
\begin{equation} \label{eq: distance cost}
    c_{\text{dist}} = \frac{d_{i+1,t}}{\alpha}, \quad i = \arg\min(\textbf{D}_t). 
\end{equation}
\normalsize

The progress cost \( c_{prog} \) evaluates the advancement of each rollout \( k \) along the path $\mathcal{P}$. Rollouts that lag behind are penalized. The index \( i_{T, k} \) of the closest waypoint at the terminal step $T$ is computed for each rollout $k$ and collected over all rollouts in the set $I_T = \{ i_{T, 1}, \dots, i_{T, K} \} $. 
Rollouts with lower waypoint indices are penalized, while those further along the path are rewarded. The progress cost is defined as:

\small
\begin{equation} \label{eq: progress cost}
    c_{\text{prog}, k} = 1 - \frac{i_{T,k} - \min(I_T)}{\max(I_T) - \min(I_T)} 
\end{equation}
\normalsize

The rotation cost \( c_{rot} \) aligns the robot’s predicted orientation \( \theta_{t} \) with the direction of the next waypoint \( \theta_{\text{target}} \). This alignment ensures smooth navigation, particularly in cluttered environments where a precise orientation is necessary for safety. The target angle \( \theta_{\text{target}} \) is computed using the arctan function, which calculates the angle between the robot’s position and the upcoming waypoint $\textbf{p}_{i+1} = [p_x, p_y]^{\top}$. The rotation cost is normalized by \( \pi \):

\small
\begin{equation} \label{eq: rotation cost}
    c_{\text{rot}} = \frac{| \theta_{\text{target}} - \theta_{t} |}{\pi}, \quad \theta_{\text{target}} = \text{arctan2}(p_y - \mathrm{x}_t^y, p_x - \mathrm{x}_t^x).
\end{equation}
\normalsize

The force cost \( c_{\text{force}} \) penalizes excessive contact forces on the robot’s joints during interactions with obstacles. IsaacGym computes the forces applied to each joint at every prediction step. Rollouts with fewer push actions accumulate less force, resulting in lower penalties over the prediction horizon \( T \). The cumulative force \( f_{\text{robot}} \) is summed across all joints $j \in N$ and prediction steps $T$, and normalized:

\small
\begin{equation} \label{eq: force cost}
    c_{\text{force}} = \frac{f_{\text{robot}}}{\max(f_{\text{robot}}) + \epsilon}, \quad f_{\text{robot}} = \sum_{t=1}^{T} \sum_{j=1}^{N} f_{j,t}.
\end{equation}
\normalsize

\subsection{Movability Evaluation}
\label{sec: movability evaluation}

The robot evaluates the movability of objects based on their perceived mass distribution in the environment. If an object initially classified as movable resists manipulation, its mass distribution is updated to the maximum threshold, reclassifying it as non-movable. 
Currently, a binary re-evaluation of movability is used to avoid excessive replanning, though it could be extended to a continuous re-evaluation for a robotic system by estimating the obstacle mass from physical data~\cite{Aguilera2021, Dutta2023}. 
Updating movability estimates and replanning are therefore initiated when specific conditions suggest that an object is non-movable.
A timer mechanism monitors these conditions and starts a timer \( t_{\text{monitor}} \) when any are triggered. If the timer exceeds \( \tau_{\text{replan}} \), the robot captures a new snapshot of the environment, updates the movability data, and regenerates the graph and path to the goal. 

The first condition for replanning is triggered when the robot’s joint velocities drop close to zero; replanning occurs if the joint velocities fall below \( \epsilon = 0.1 \, \text{m/s} \), indicating significant resistance. The second condition involves a deviation between actual and desired joint velocities, with a deviation greater than 75\% (\( \lambda = 0.75 \)) of the desired velocity signaling ineffective movement. Lastly, if the joint velocities are high but the robot's position remains stable, indicating slipping, this condition is met when the robot is stationary but joint velocities exceed \( \mu = 0.1 \, \text{m/s} \). If none of these conditions are met, the timer resets, pausing the evaluation process until one of the conditions is activated again.

\section{Experiments and Results}
\label{section:experiments_and_results}

This section outlines the experimental setup and results, validating the proposed SVG-MPPI algorithm in both simulated and real-world environments. The algorithm's performance is evaluated based on its ability to navigate toward obstructed goal positions, and we compare it against other established methods.

\subsection{Experimental Framework}
The experiments are conducted using a Dingo-O robot, a holonomic, velocity-controlled platform developed by Clearpath Robotics. In real-world trials, precise localization of the robot and obstacles is achieved through a Vicon motion capture system, with poses updated in Isaac-Gym at 25 Hz to ensure accurate MPPI rollouts. A maximum pushable mass of 30 kg is enforced, along with a 0.3-meter safety margin based on the robot’s width (517 mm) to prevent collisions. The waypoint interpolation interval is set to 0.5 meters, striking a balance between computational efficiency and path smoothness. Replanning is triggered if the conditions outlined in Section~\ref{sec: movability evaluation} persist for more than 30 seconds. MPPI control operates with a time step of 0.08 seconds and a prediction horizon of 25 steps, covering 2 seconds of movement. Further experimental details are available in our \href{https://github.com/tud-amr/SVG-MPPI}{GitHub repository}.

\begin{figure}[tb!]
    \centering
    \begin{minipage}{0.49\textwidth}
        \includegraphics[width=\textwidth]{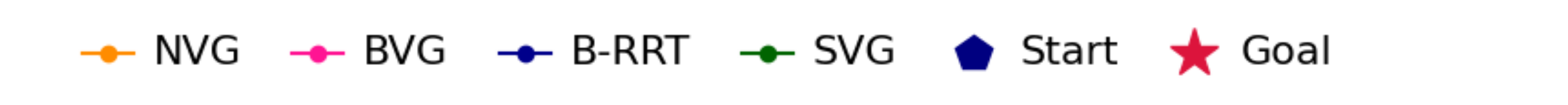}
    \end{minipage}\hfill
    \begin{minipage}{0.1\textwidth}
        \centering
        \includegraphics[width=\textwidth]{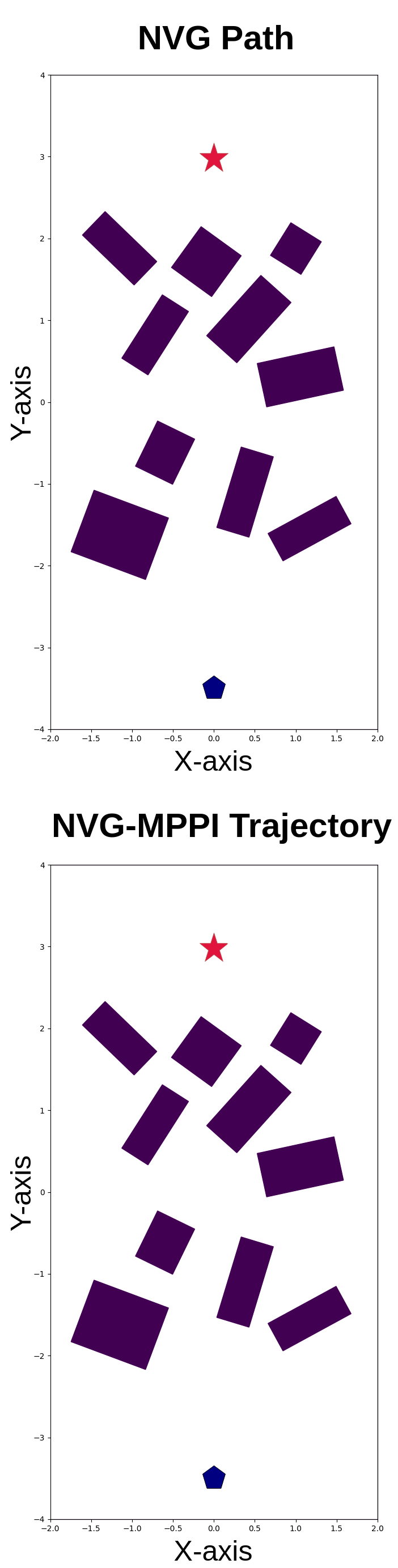}
        \caption*{\hspace{7mm} (a)}
    \end{minipage}\hfill
    \begin{minipage}{0.1\textwidth}
        \centering
        \includegraphics[width=\textwidth]{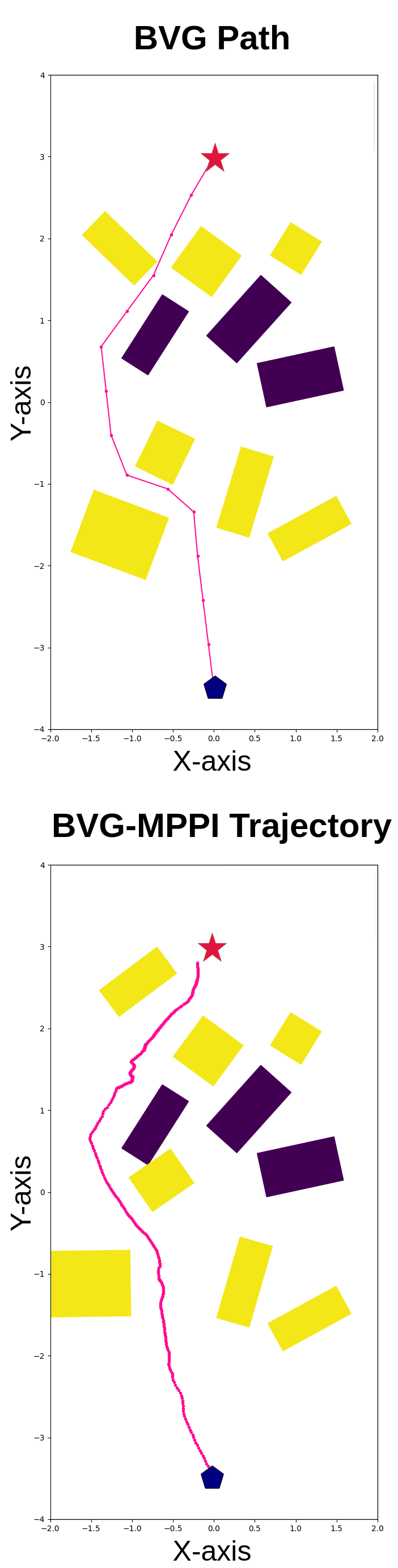}
        \caption*{\hspace{7mm} (b)}
    \end{minipage}\hfill
    \begin{minipage}{0.1\textwidth}
        \centering
        \includegraphics[width=\textwidth]{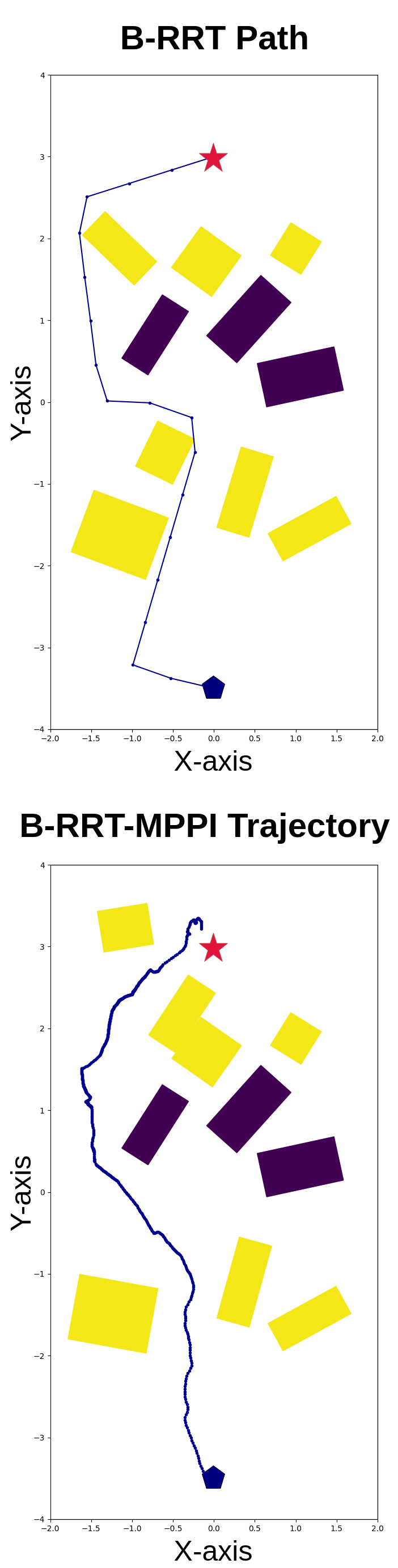}
        \caption*{\hspace{7mm} (c)}
    \end{minipage}\hfill
    \begin{minipage}{0.1\textwidth}
        \centering
        \includegraphics[width=\textwidth]{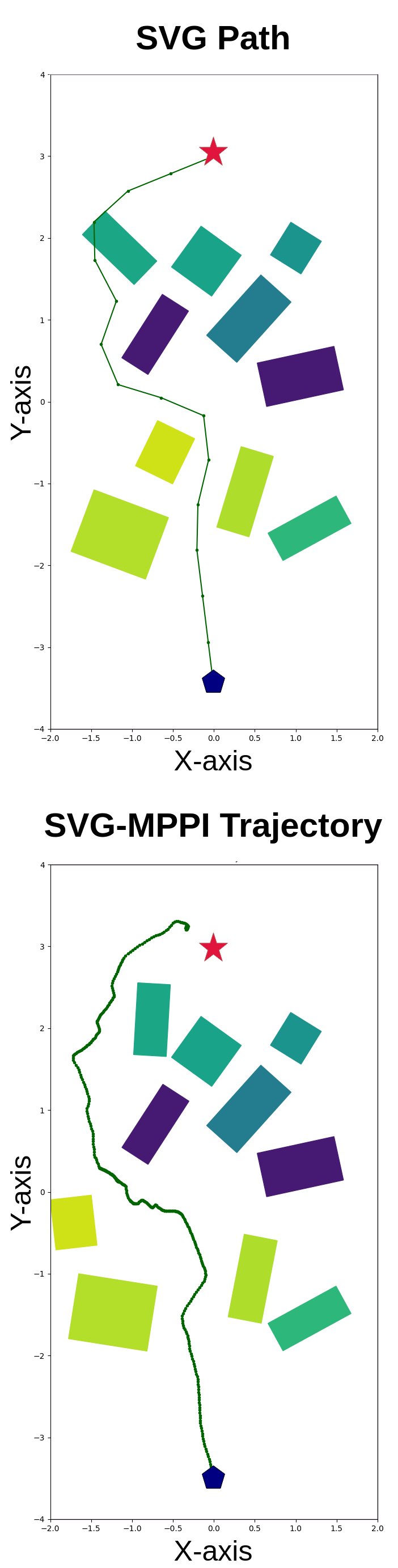}
        \caption*{\hspace{7mm} (d)}
    \end{minipage}\hfill
    \begin{minipage}{0.051\textwidth}
        \raisebox{5mm}{\includegraphics[width=\textwidth]{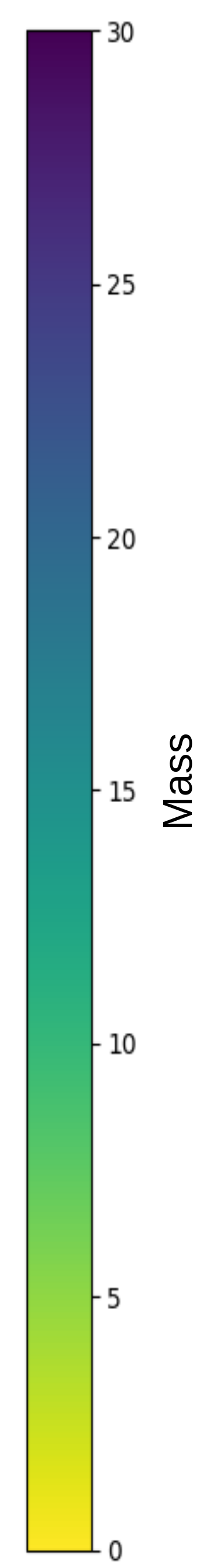}}
    \end{minipage}
    \caption{Comparison of the different path planning strategies: (a) NVG, (b) BVG, (c) B-RRT, and (d) SVG. Each figure shows the planned path and the robot’s trajectory using MPPI as the local motion planner. For NVG, no path could be found as all obstacles are considered non-movable.}
    \label{fig:path_planning_strategies_per_planner}
    \vspace{-4mm}
\end{figure}

We evaluate the algorithm in a simulated 4x8 meter cluttered room, where 20\% of the space is occupied by movable obstacles and 5\% by stationary obstacles where obstacles are randomized with respect to their positions, dimensions and masses (see Fig.~\ref{fig:path_planning_strategies_per_planner}). This randomly structured environment challenges the robot to navigate tight spaces, where pushing is often required if avoidance is unfeasible. The randomness ensures robust testing by simulating real-world conditions, making it an effective benchmark for adaptability, efficiency, and real-time replanning capabilities.

\subsection{Benchmark Algorithms} \label{sec: benchmarks}
To the best of our knowledge, no existing algorithm directly mirrors the SVG-MPPI approach. Therefore, we developed three comparative methods to benchmark our system, focusing on how obstacle movability is interpreted in both global path planning and local control strategies.

The first method, NVG (Normal Visibility Graph), treats all obstacles as stationary. The second, BVG (Binary Visibility Graph), uses a binary classification, distinguishing between movable and immovable obstacles. The third method, B-RRT (Binary Rapidly-exploring Random Tree), also applies a binary classification of movability but uses an RRT-based path planning algorithm. In this method, sampling within the safety margin of obstacles is permitted if their mass is below 30 kg, though no additional node costs are applied. The RRT approach follows the traditional structure as outlined in \cite{Lavalle1998}. Fig.~\ref{fig:path_planning_strategies_per_planner}a-d illustrates how each planner interprets movability and demonstrates a sample scenario where a path is calculated from the starting position to the goal.

In addition to comparing global path planning methods, we assess the impact of different movability interpretations on local control strategies by pairing each path planning method with MPPI. NVG-MPPI does not account for movability and treats all obstacles as stationary, while BVG-MPPI and B-RRT-MPPI apply a binary movability distinction in both the path and trajectory planner. In contrast, SVG-MPPI employs a continuous approach to movability. This comparison allows us to evaluate how different interpretations of movability affect both path planning and control performance.

\subsection{Performance Metrics}
Performance metrics were collected over 50 simulation runs, with obstacle masses randomly assigned between 4 and 36 kg. The path planners, NVG, BVG, B-RRT and SVG, are compared on their \textit{Path Planning Success} indicating the percentage of paths found from start to goal and \textit{Planner time} expressing the average computation time to generate the path. All planners are combined with MPPI as local motion planning method and analyzed on their \textit{Execution Success} providing the percentage of scenarios where the simulated robot reaches the goal and its corresponding average \textit{Execution time}. In addition, we include the cumulative force averaged over the 50 randomly generated scenarios to analyze the contact forces between the robot and obstacles over all successful trials. 

\begin{table*}[ht]
    \caption{\small{Experimental results of the compared methods for 50 scenarios with objects of randomized size, placement, and mass.}}
    \label{table:combined}
    
    \renewcommand{\arraystretch}{1.2}
    \resizebox{\textwidth}{!}{
    \begin{tabular}{!{\color{lightgray}\vrule width 2pt}>{\columncolor{lightgray}}c c c}
        \toprule
        \textbf{Path} & \textbf{Path Planning} & \textbf{Planner Time (s)} \\
        \textbf{Planner} & \textbf{Success (\%)}& \textbf{(Mean $\pm$ SE)} \\
        \midrule
        NVG  & 16.4 & 0.169 $\pm$ 0.014    \\ 
        BVG  & 98.2  & 0.195 $\pm$ 0.006  \\ 
        B-RRT & 92.7 & 1.120 $\pm$ 0.483  \\ 
        \textbf{SVG}  & \textbf{98.2} & \textbf{0.196 $\pm$ 0.006} \\ 
        \bottomrule
    \end{tabular}
    
    \hspace{1mm}

    \begin{tabular}{!{\color{lightgray}\vrule width 2pt}>{\columncolor{lightgray}}c c c c}
        \toprule
        \textbf{Control} & \textbf{Execution to Goal} & \textbf{Execution Time (s)} & \textbf{Cumulative Force (kN)} \\
        \textbf{Strategy} &  \textbf{Success (\%)} & \textbf{(Mean $\pm$ SE)} & \textbf{(Mean $\pm$ SE)} \\
        \midrule
         NVG-MPPI & 12.7  & 109 $\pm$ 18 & 39 $\pm$ 23 \\ 
         BVG-MPPI & 49.1  & 129 $\pm$ 13 & 165 $\pm$ 418 \\ 
         RRT-MPPI & 50.9 & 130 $\pm$ \ 9 & 175 $\pm$ 31   \\ 
         \textbf{SVG-MPPI} & \textbf{54.5}  & \textbf{120 $\pm$ 11} & \textbf{97 $\pm$ 17} \\ 
        \bottomrule
    \end{tabular}
    }
\vspace{-3mm}
\end{table*}
\normalsize

\subsection{Experimental analysis}
\label{sec: experimental analysis}

In Table~\ref{table:combined}, the performance metrics for SVG-MPPI are compared against benchmark planners as introduced in Sec.~\ref{sec: benchmarks} across 50 randomized scenarios. The proposed path planner SVG consistently outperforms NVG and B-RRT in path planning success rates with 98.2\% for SVG and 16.4\% and 92.7\% for NVG and B-RRT respectively.  
By integrating the continuous movability within the path planner SVG and the local planner MPPI, we obtain an improved execution success rate, 54.5\%, compared to a binary classification method, 49.1\% and 50.9\% for BVG and B-RRT respectively, with a mildly improved execution time. 
Most importantly, SVG-MPPI also demonstrates significantly lower cumulative force compared to B-RRT-MPPI and BVG-MPPI, emphasizing its efficiency in minimizing push actions during navigation. Its strategic node placement, effort-based path selection, and considered continuous movability in both the local and global planner, lead to smoother interactions with the environment, making it particularly effective in scenarios requiring multiple obstacle manipulations. 
Note that NVG-MPPI can only solve scenarios where a collision-free path can be constructed. The minimal cumulative force in Table~\ref{table:combined} results from unintended pushing due to suboptimal turns and path-following.

An illustrative example can be observed in Fig.\ref{fig:path_planning_strategies_per_planner} expressing the paths and trajectories of all compared methods. By considering continuous movability, SVG-MMPI selects a path near the lighter obstacles that lead to a small displacement of the obstacles, e.g. a low push effort over the trajectory. B-RRT-MPPI, on the other hand, pushes one obstacle across the room, and both BVG-MPPI and B-RRT-MPPI push one obstacle against another obstacle resulting in additional push effort. NVG-MPPI is unable to generate a path in this environment as all obstacles are considered non-movable. SVG-MPPI applied replanning three times without success, but in two cases, it enabled the robot to reach the goal. Other methods did not include replanning.
Moreover, SVG-MPPI’s planner time remains highly efficient, 0.196 $\pm$ 0.006, supporting real-time replanning in dynamic environments. This combination of efficiency and adaptability highlights SVG-MPPI’s robustness, particularly in environments with high obstacle density and frequent replanning needs.

\begin{figure}[tb!]
    \centering
    \begin{minipage}{0.42\textwidth}
        \centering
        \includegraphics[width=\textwidth]{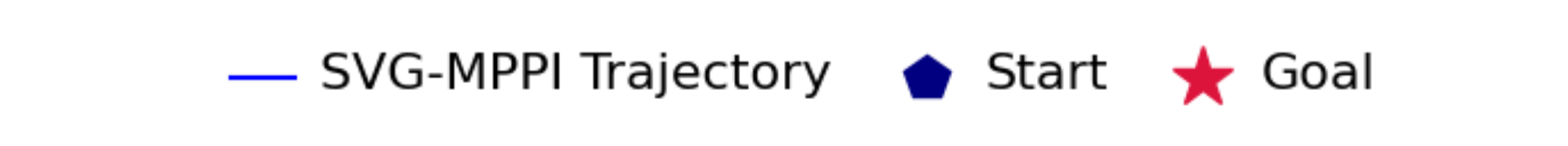}
    \end{minipage}\hfill
    \begin{minipage}{0.23\textwidth}
        \centering
        \includegraphics[width=\textwidth]{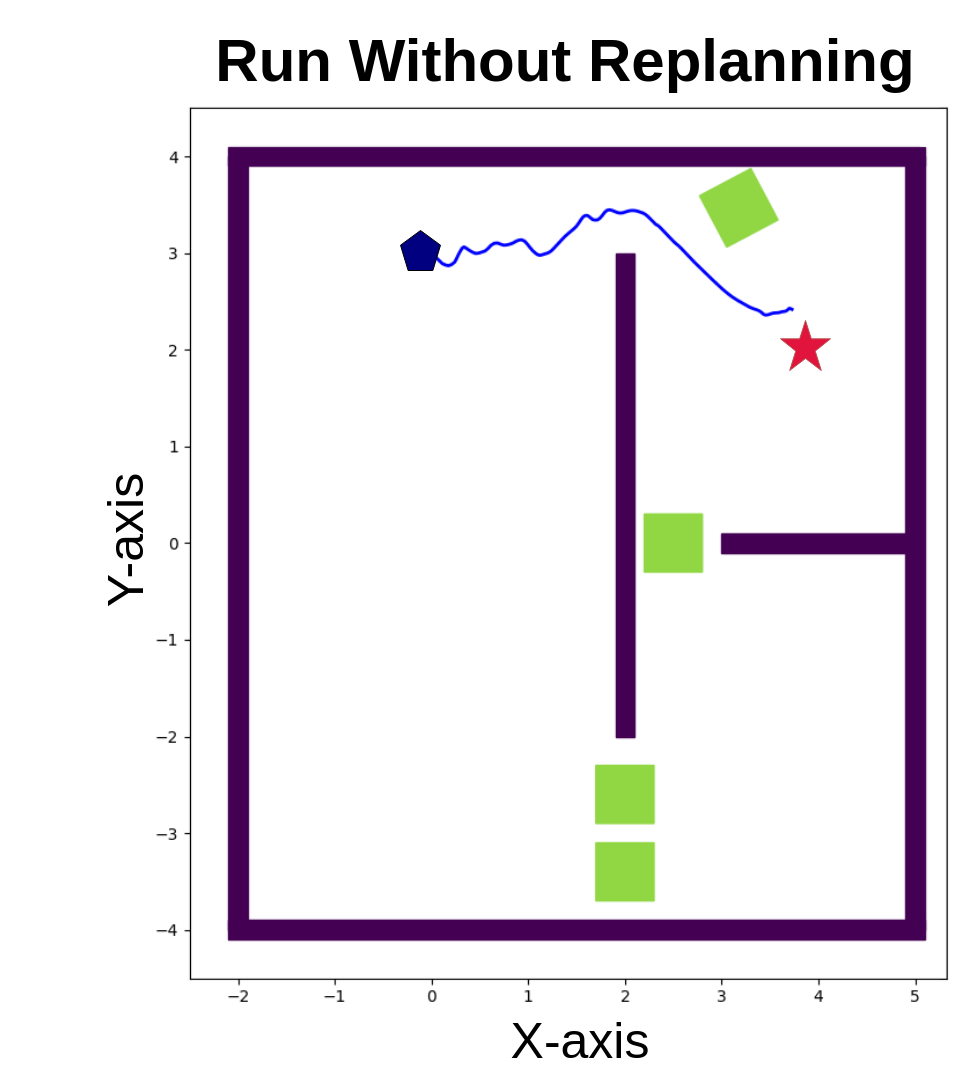}
        \caption*{\hspace{22mm} (a)}
    \end{minipage}\hfill
    \begin{minipage}{0.23\textwidth}
        \centering
        \includegraphics[width=\textwidth]{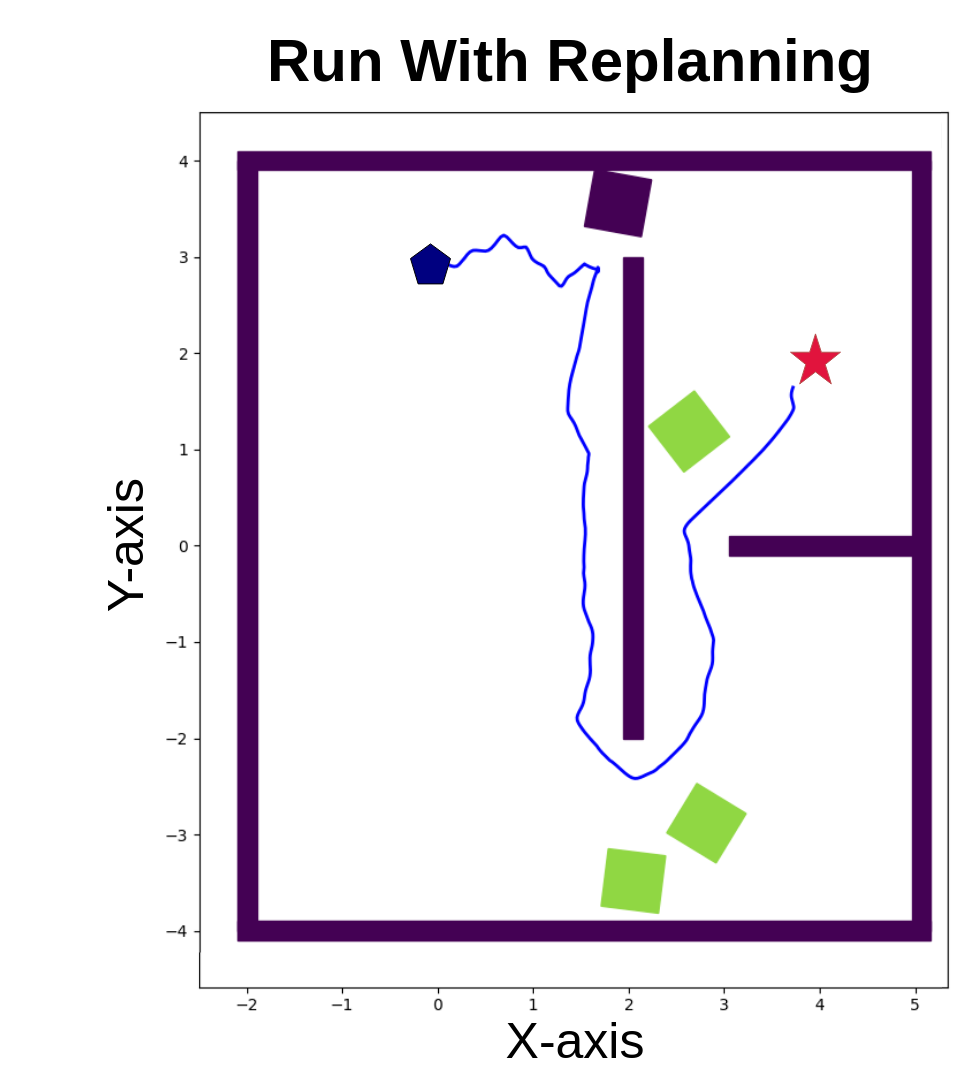}
        \caption*{\hspace{22mm} (b)}
    \end{minipage}
    \caption{Comparison between (a) the trajectory without replanning, where the robot moves a single obstacle to create a path, and (b) the trajectory with replanning, triggered when the estimated movability is updated and the robot reroutes through another path.}
    \label{fig:movability_evaluation}
    \vspace{-3mm}
\end{figure}

\subsubsection{Online Replanning} \label{sec: replanning}
Fig.~\ref{fig:movability_evaluation} demonstrates two scenarios where in scenario (a) the obstacle mass is as expected and in scenario (b) the obstacle is observed to be non-movable when encountered. 
In scenario (a), the robot moves a single obstacle to create a pathway along the upper side of the room. In scenario (b), the robot is unable to move the obstacle which triggers the replanning process as described in Sec.~\ref{sec: movability evaluation}. 
This allowed the robot to adapt by rerouting through the lower side of the room, demonstrating the ability of SVG-MPPI to update movability online and replan accordingly.

\subsubsection{Real-World Experiments}
In the real-world test, the Dingo-O robot navigates a hallway with movable and immovable obstacles to confirm the practical effectiveness of the SVG-MPPI algorithm. The robot successfully created a passage by relocating obstacles, reaching the goal despite the physical constraints of the environment, as shown in Fig.~\ref{fig:overview_of_svg_mppi} and the accompanying video. During the real-world tests, the robot encounters three movable obstacles where the lightest obstacle, i.e. obstacle C, is moved forward and the robot navigates in the created free space between obstacles B and C. During the real-world experiments, obstacle B is slightly pushed due to model mismatch and additional disturbances in the real-world tests compared to the simulated scenario.

\section{CONCLUSION} 
\label{section: conclusion}

Integrating obstacle movability into robot navigation enables access to previously unreachable areas through obstacle manipulation. We introduce SVG-MPPI, a framework for navigating environments with movable obstacles that minimizes push actions and avoids explicit obstacle placement. SVG informs MPPI by strategically placing nodes around movable obstacles using a continuous movability cost. Instead of relying on explicit contact and dynamics models, SVG-MPPI leverages the Isaac-Gym physics engine for real-time state transitions and contact force data to reduce push actions. Our proposed method outperforms other algorithms, including NVG-MPPI with no movability information, BVG-MPPI and B-RRT-MPPI with a binary notion of movability, in terms of success rate and contact force minimization. Future work could incorporate friction into the movability factor and introduce gain scheduling for more adaptive behavior in extreme movability cases.



\printbibliography[heading=bibintoc,title=References]

\end{document}